\documentclass[conference]{IEEEtran}

\usepackage{graphicx}
\usepackage{comment}
\usepackage{amsmath,amssymb} 
\usepackage{color}
\usepackage{url}
\usepackage{hyperref}
\usepackage{cite}
\usepackage{algorithmic}
\usepackage{algorithm}
\usepackage{soul}

\ifCLASSINFOpdf
\else
\fi

\begin{document}
%
\title{A Unified DNN Weight Compression Framework Using Reweighted Optimization Methods}

\author{Tianyun Zhang$^{1,4}$,
        Xiaolong Ma$^{2,4}$,
        Zheng Zhan$^{2}$,
        Shanglin Zhou$^{3}$,
        Minghai Qin$^{4}$,
        Fei Sun$^{4}$,
        \\Yen-Kuang Chen$^{4}$,
        Caiwen Ding$^{3}$,
        Makan Fardad$^{1}$,
        Yanzhi Wang$^{2}$,
\\1. Syracuse University 2. Northeastern University  \\ 3. 	University of Connecticut 4. Alibaba Group   }


%


\maketitle

\begin{abstract}

To address the large model size and intensive computation requirement of deep neural networks (DNNs), weight pruning techniques have been proposed and generally fall into two categories, i.e., static regularization-based pruning and dynamic regularization-based pruning. However, the former method currently suffers either complex workloads or accuracy degradation, while the latter one takes a long time to tune  the  parameters to  achieve  the  desired  pruning  rate without accuracy loss. In this paper, we propose a unified DNN weight pruning framework with dynamically updated regularization terms bounded by the designated constraint, which can generate both non-structured sparsity and different kinds of structured sparsity. We also extend our method to an integrated framework for the combination of different DNN compression tasks.



\end{abstract}


\section{Introduction}


DNNs have achieved impressive results in many fields including image classification \cite{krizhevsky2012imagenet,he2016deep}, natural language processing \cite{hinton2012deep,dahl2012context}, self-driving cars \cite{makantasis2015deep}, etc. The state-of-the-art DNNs have large model size and computational requirement, which impedes the critical requirements (e.g., real time, low power) in the inference phase. To address these challenges, prior works have focused on developing DNN model compression techniques, such as weight pruning \cite{han2015learning,wen2016learning,guo2016dynamic,mao2017exploring,zhang2018systematic} and weight quantization \cite{han2016deep_compression,park2017weighted,leng2018extremely,zhou2017incremental,ren2019admm}. 

Weight pruning has gained more popularity since it offers more potential pruning rate~\cite{han2016deep_compression}. The objective of DNN weight pruning is to reduce the number of non-zero elements in the weight matrix while maintaining the prediction accuracy. Early works in weight pruning utilize a static, magnitude-based method \cite{han2015learning} or $\ell_{1}$-based regularization \cite{wen2016learning} to explore sparsity in DNN models.
Although these methods can prune the weights without accuracy loss,
it is heuristic and can only find {a small parts of non-critical weights to prune}.  
To overcome the accuracy degradation while further prune the DNNs, several works \cite{li2017pruning,luo2017thinet,zhuang2018discrimination,zhang2018adam} propose more well-developed methods to increase the pruning rate based on the sparsity types proposed by \cite{wen2016learning}. 
Recently, a work \cite{zhang2018systematic} uses the alternating direction method of multiplier (ADMM)~\cite{boyd2011distributed} to solve the $\ell_{0}$ constraint problem and achieve good performance on pruning rate without accuracy loss. With the powerful ADMM optimization framework, pruning problems are re-formed into optimization problems with the dynamically updated regularization terms bounded by the designated hard constraint sets which enable arbitrary desired pruning dimensions to fulfill the vast design space. 
However, ADMM suffers from long convergence time due to the strong non-convexity of the $\ell_{0}$ constraints. Additionally, it is a highly time-consuming process to set the hyperparameters manually in a hard constraints problem, which intrinsically is a heuristic exploration that mainly relies on the experiences of the designer, and the derived hyperparameters are typically sub-optimal. It is imperative to find a better solution to solve the pruning problem with high efficient and self-adaptive regularization that can automatically determine pruning hyperparameters and maintain accuracy.

In this paper, we propose a unified DNN weight pruning framework with dynamically updated regularization terms bounded by the designated constraint, which can generate both non-structured sparsity and different kinds of structured sparsity. 
In non-structured pruning, we need to reduce the $\ell_{0}$ norm of the weight matrix, but it is an intractable problem since $\ell_{0}$ norm is non-convex and discrete. To deal with this issue, we solve the reweighted $\ell_{1}$ problem \cite{candes2008enhancing} as the proxy of $\ell_{0}$ problem. 
Structured pruning requires not only the sparsity in weights but also the position of the zeros elements. 
To generate different kinds of group sparsity, 
we propose to use a reweighted method on group lasso regularization. 
In our proposed framework, we first use a reweighted method to regularize the model, then remove the weights which are close to zero and mask the gradient of these weights to ensure that they no longer update. We retrain the remaining non-zero weights to retrieve the accuracy. 
We adopt the reweighted regularization method with designated sparsity types, which avoids strongly non-convex $\ell_{0}$-norm based hard constraints in the state-of-the-art ADMM formulation, therefore accelerates the convergence and reduces the number of hyperparameters. 
Since the loss function of DNNs is non-convex, when we use the reweighted method we also need to solve a non-convex problem and cannot achieve the globally optimal solution. This motivates us to use the reweighted method again on the sparse model achieved in our first step. After implementing the reweighted method for several more steps, we achieve higher pruning rate. 

We also extend our proposed method to an integrated framework for the combination of different DNN weight compression tasks. In this framework we use ADMM to deal with one compression task with hard constraints and apply the reweighted method to search for critical weights or structures in the other compression tasks. This achieves significant improvement compared with applying the ADMM-based hard constraints on both of the tasks.

Our major contributions in this paper can be summarized as follows:
\begin{itemize}

\item We propose a unified DNN weight pruning framework with the reweighted regularization technique, which achieves noticeable improvement on pruning rate compared with the magnitude-based, $\ell_1$-based regularization and ADMM-based hard constraint methods. 
As an example, our methods achieve 4.2$\times$ structured pruning rate with 88.5\% top-5 accuracy on ResNet-18 for ImageNet while the state-of-the-art ADMM methods achieve 3.0$\times$ structured pruning rate with 87.9\% top-5 accuracy. 

\item We adopt a dynamically updated regularization, which avoids the strong non-convexity from the $\ell_{0}$-norm based hard constraints in the state-of-the-art ADMM formulation that causes excessive convergence time and complex hyperparameter settings.
Thereby we can use this automatic scheme to derive the per-layer pruning rate that results in outperforming pruning results. 
The training time can be reduced from 150 epochs (ADMM-based methods) to 85 epochs using our proposed methods for a single set of hyperparameters. The ADMM-based hard constraint method need to set per-layer pruning rates, which leads to a large amount of hyperparameters, while our proposed methods only needs set one single penalty parameter and the per-layer pruning rates can be determined automatically.


\end{itemize}




\section{Related Works}
\label{sec:related}

\subsection{Weight pruning} 

\subsubsection{Non-structured and structured pruning}
Early works in weight pruning utilize a static, magnitude-based method or $\ell_{1}$-based regularization to explore sparsity in DNN models~\cite{han2015learning,wen2016learning,he2017channel}. With specified regularization dimensions on weight vectors, we can perform different types of  pruning method including \emph{non-structured pruning} and \emph{structured pruning}, but with limited pruning rates and non-negligible accuracy degradation  due to the intrinsically heuristic and non-optimized approach. Non-structured pruning in~\cite{han2015learning} iteratively prunes weights at arbitrary location based on their magnitude, resulting in a sparse model to be stored in the compressed sparse column (CSC) format. 
It leads to an undermined processing throughput because the indices in the compressed weight representation cause stall or complex workload on highly parallel architectures.  
Structured pruning incorporates regularity in weight pruning, including filter-wise pruning and shape-wise pruning~\cite{wen2016learning,he2017channel}, targeting at generating regular and smaller weight matrices based on $\ell_{1}$ regularization to eliminate overhead of weight indices and achieve higher acceleration. 
The weight matrix will maintain a full matrix but with reduced dimensions, and indices are no longer needed. 
\subsubsection{Pattern and kernel-wise pruning}
Recently, special dimensions pruning has been studied in~\cite{chen2018sc} that the sparse complimentary kernels can save half of the weights and computations, but it only focuses on parameter and computation reduction without discussing on platform acceleration. 
Inspired by the prior work of ADMM, another kernel level pruning in~\cite{ma2019pconv} investigates kernel-wise pruning in pattern-based sparsity which utilizes hard constraints involved in the dynamic regularization term to facilitate its special pruning dimension. With the supports from compiler-assisted inference framework, it can achieve platform acceleration at edge computing devices such as mobile platforms.
\subsection{Weight Quantization} 
Weight quantization has been investigated as an orthogonal model compression technique~\cite{lin2016fixed,wu2016mobile,li2016ternary,courbariaux2015binaryconnect}. Various quantization techniques have demonstrated to achieve weight precision with tolerable accuracy loss on different DNN models including fixed bit-length, ternary and even binary weight representations~\cite{li2016ternary,courbariaux2015binaryconnect}. Weight quantization can simultaneously reduce the DNN model size, computation and memory access intensity and is naturally hardware-friendly, 
since both storage and computation of DNNs will be reduced proportionally.
Other prior works have investigated the combination of weight quantization and weight pruning~\cite{han2015deep,ren2019ADMMNN}. Since they leverage different sources of redundancy, they may be combined to achieve higher DNN compression ratio.

\section{A Unified Framework of DNN Pruning}
\label{sec:framework}

\subsection{Motivation}

We observe that many early works use static methods on the weight pruning of DNNs, e.g. the magnitude based methods in \cite{han2015learning} and the $\ell_{1}$ regularization method in \cite{wen2016learning}. We propose two hypotheses based on these methods. First, some weights with small magnitude are critical to maintain the accuracy of the model, thus we cannot prune the weights simply based on their magnitude. Second, the $\ell_{1}$ norm is not a good approximation of the $\ell_{0}$ norm, and using the $\ell_{1}$ regularization will penalize some critical weights to values close to zero. These two hypotheses motivate us to find a better approximation for the $\ell_{0}$ norm in order to generate highly sparse model without accuracy degradation.

\begin{figure*}[ht]
\centering
\includegraphics[width=0.99\linewidth]{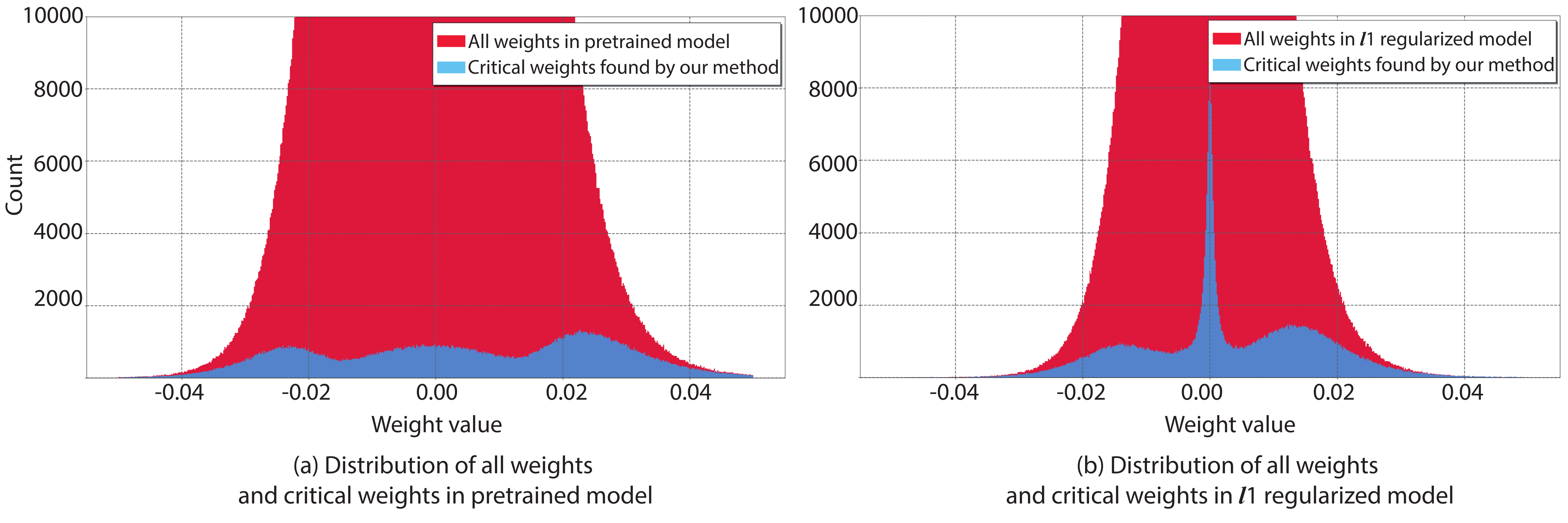}  
\caption{Comparisons of the distribution of all weights and critical weights (remaining weights after pruning) in the second fully connected layer (FC-2) of a pretrained AlexNet model and $\ell_{1}$ regularized model.}
\label{fig:distribution}
\end{figure*}

We prune an AlexNet using the reweighted method to verify our two hypotheses. Fig.~\ref{fig:distribution} shows the histogram of the weights in the second fully connected layer (FC-2) of AlexNet. In Fig.~\ref{fig:distribution} (a), the red area is the histogram of the original weight distribution without pruning (we omit the top part of the distribution due to space limitations), and the blue area is the distribution after removing 97.9\% of the weights by the reweighted method without accuracy loss. We can observe that the critical weights (remaining weights after pruning) are approximately uniformly distributed, which means that some weights with small magnitude are also critical.
This verifies our first hypothesis. In Fig.~\ref{fig:distribution} (b), the red area is the histogram of the weights after regularizing the FC-2 layer using $\ell_{1}$ regularization. Comparing with the red area of Fig.~\ref{fig:distribution} (a), $\ell_{1}$ regularization reduces the magnitude of the weights in the entire network. The critical weights, shown in the blue area of Fig.~\ref{fig:distribution} (a), have a different distribution after $\ell_{1}$ regularization is applied, as shown in the blue area of Fig.~\ref{fig:distribution} (b). It is clear that $\ell_{1}$ regularization penalizes a lot more critical weights towards zero, as is  shown in the high peak of the blue area. After pruning, those critical weights will be forced to zero, and thus may negatively impact the model quality. 
This is because $\ell_{1}$ regularization casts equal penalty on all weights or weight groups. This violates the original intention of weight pruning, which is to remove the ``non-critical" weights instead of ``small" weights, thus it is not a good approximation for $\ell_{0}$ regularization, which verifies our second hypothesis.

Besides static methods used in the early works, a recent work \cite{zhang2018systematic} focuses on $\ell_{0}$ norm based optimization with an ADMM-based hard constraint approach and achieves good performance on DNN pruning without accuracy loss. This method first formulates weight pruning as an optimization problem with a hard constraint on $\ell_{0}$ norm, and then uses ADMM~\cite{boyd2011distributed} to solve the problem. 
In the ADMM-based solution framework, the regularization term is dynamically updated in each iteration, and it achieves better performance compared with the work based on static methods. 
However, because of the hard constraint on the $\ell_{0}$ norm, the degree of sparsity in each DNN layer needs to be pre-specified. 
This fact limits the flexibility of the ADMM-based method. 
In reality, when the degree of sparsity undefined, it is hard to determine the numbers of weights to prune for each layer.
Therefore, ADMM-based method may take an excessive amount of time to tune the parameters to achieve the desired pruning rate without accuracy loss.

\begin{figure*}[t]
\centering
\includegraphics[width=1\linewidth]{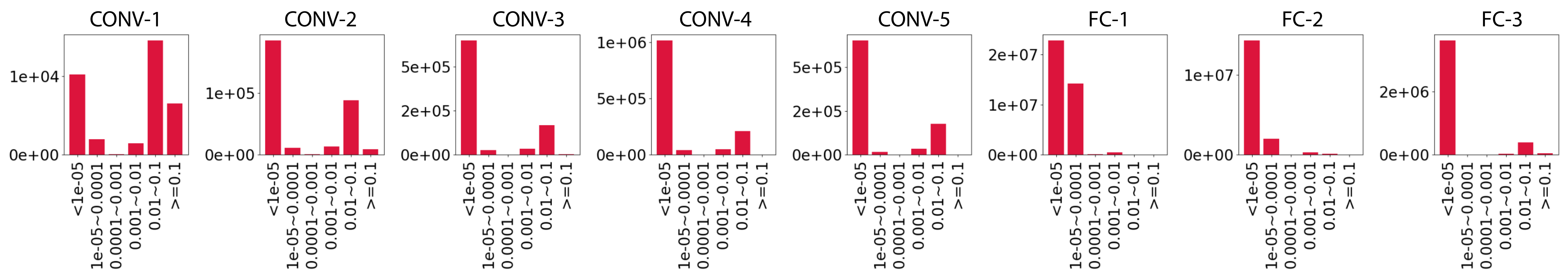}  
\caption{Weight distribution of all layers in AlexNet after our reweighted $\ell_1$ regularization method}
\label{fig:th_distribution}
\end{figure*}

In this paper, we propose to use a reweighted method for DNN weight pruning. It is a dynamic regularization-based method, where in each iteration the penalties on different weights are dynamically updated. Different from the ADMM-based method in which the hyperparameters need to be tuned, we only need to 
set a single penalty parameter in our method, the value of this parameter is easy to set and we will discuss it in Section~\ref{Sec_exp}.After training with our reweighted regularization method, we can decide the degree of sparsity in each layer based on the distribution of weights. For example, the weight distribution of each layer in AlexNet after our reweighted $\ell_1$ regularization method is shown in Fig.~\ref{fig:th_distribution}. We can observe that most of the weights with large magnitude are distributed in the range of $0.01$ to $0.1$, and most of the weights with small magnitude are smaller than $0.0001$. This means small weights are 100$\times$ or more smaller than large weights. Thus, removing the weights with magnitude smaller than $0.0001$ has a minor effect on the accuracy of the DNN. Meanwhile, we do not need to specify the pruning rate for each layer as it will be determined dynamically by our proposed reweighted regularization method.


\subsection{Non-structured Pruning}

Consider an $N$-layer DNN, the collection of weights and biases of the $i$-th layer is denoted by $W_i$ and $b_i$, respectively. The loss function associated with the DNN is denoted by $f \big( \{{{W}}_{i}\}_{i=1}^N, \{{{b}}_{i} \}_{i=1}^N \big)$. In DNN training, we minimize the loss function to increase the accuracy. For the non-structured weight pruning problem, our aim is to reduce the number of non-zero elements in the weight matrix while maintaining the accuracy. Therefore, we need to
minimize the summation of the loss function and the $\ell_{0}$ regularization term as follows,
\begin{equation*}
 \underset{ \{{{W}}_{i}\},\{{{b}}_{i} \}}{\text{minimize}}
\ \ \ f \big( \{{{W}}_{i} \}_{i=1}^N, \{{{b}}_{i} \}_{i=1}^N \big)+\lambda\sum_{i=1}^{N} \| W_i \|_{\ell_0}, \\
\end{equation*}
where $\lambda$ is the penalty parameter to adjust the relative importance of accuracy and sparsity.

The problem with the $\ell_{0}$ norm is intractable, thus we use a reweighted $\ell_{1}$ method \cite{candes2008enhancing} to approximate the $\ell_{0}$ norm. 
For the reweighted $\ell_{1}$ method, we instead solve the problem
\begin{equation}
\label{prob_general}
 \underset{ \{{{W}}_{i}\},\{{{b}}_{i} \}}{\text{minimize}}
\ \ \ f \big( \{{{W}}_{i} \}_{i=1}^N, \{{{b}}_{i} \}_{i=1}^N \big)+\lambda\sum_{i=1}^{N} R({{P}}_{i}^{(l)},{{W}}_{i}), \\
\end{equation}
where $ R({{P}}_{i}^{(l)},{{W}}_{i}) = \| P_{i}^{(l)} \circ W_i \|_{\ell_1}$, the operator $\circ$ denotes element-wise multiplication, and $P_{i}^{(l)}$ is the collection of penalties on different weights, which is updated in every iteration to increase the degree of sparsity beyond the $\ell_{1}$ norm regularization. In each iteration, we denote the solution of ${{W}}_{i}$ by ${{W}}_{i}^{(l)}$ and update $P_{i}$ by setting $P_{i}^{(l+1)} = \frac{1}{{|{W}}_{i}^{(l)}|+\epsilon}$,
where $|\cdot|$ denotes the absolute value and $\epsilon$ is a small parameter to avoid dividing by zero. In our experiment $\epsilon=0.001$ works well. All operations above are performed element-wise.

\subsection{Structured Pruning}

Filter-wise pruning, shape-wise pruning and kernel-wise pruning are a subset of structured pruning. Different from non-structured pruning, structured pruning requires not only the sparsity in weights but also the position of the zeros elements \cite{wen2016learning}. To generate different kinds of group sparsity, we propose to use the reweighted method on the group lasso regularization \cite{yuan2006model}. Problem (\ref{prob_general}) is also applicable to structured pruning. For filter-wise pruning, the regularization term is 
\[
R({{P}}_{i}^{(l)},{{W}}_{i}) = \sum_{a=1}^{A} \| P_{i,a}^{(l)} \circ  (W_i)_{a,:,:,:} \|_F^2,
\]
where $(W_i)_{a,:,:,:}$ denotes the $a$-th filter of $W_i$, and $P_{i,a}$ is updated by
$
P_{i,a}^{(l+1)} = \frac{1}{\|{(W_i)_{a,:,:,:}^{l} \|_F^2+\epsilon}}.
$

For shape-wise pruning, the regularization term is \[R({{P}}_{i}^{(l)},{{W}}_{i}) = \sum_{b=1}^{B} \sum_{c=1}^{C} \sum_{d=1}^{D} \| P_{i,b,c,d}^{(l)} \circ  (W_i)_{:,b,c,d} \|_F^2,
\]
where $(W_i)_{:,b,c,d}$ denotes the collection of weights located at position $(:, b, c, d)$ in every filter and $P_{i}$ is updated by
$
P_{i,b,c,d}^{(l+1)} = \frac{1}{\|{(W_i)_{:,b,c,d}^{l} \|_F^2+\epsilon}}.
$

For kernel-wise pruning, the regularization term is
\[
R({{P}}_{i}^{(l)},{{W}}_{i}) = \sum_{a=1}^{A} \sum_{b=1}^{B} \| P_{i,a,b}^{(l)} \circ  (W_i)_{a,b,:,:} \|_F^2,
\] 
and $P_{i}$ is updated by
$
P_{i,a,b}^{(l+1)} = \frac{1}{\|{(W_i)_{a,b,:,:}^{l} \|_F^2+\epsilon}}.
$

\subsection{A Unified Algorithm for Non-structured and Structured Sparsity}
\begin{algorithm}[tb]
  \caption{Unified reweighted
  method on DNN pruning}
  \label{alg:example2}
\begin{algorithmic}
  \STATE {\bfseries Input:} pretrained model
  \STATE Initialize $P_{i}$
  \STATE Set $l = 1$
  \STATE Set $T$ as the number of iterations of the reweighted method
  \FOR{$l \le T$ }
        \STATE Solve the regularization Problem (\ref{prob_general}) using SGD or ADAM
        \STATE Update $P_{i}^{(l+1)}$ using the solution of $W_{i}^{(l)}$
  \ENDFOR
  \STATE Remove the weights (or group of weights) which are close to zeros and retrain the DNN using the non-zero weights
\end{algorithmic}
\end{algorithm}

In \cite{candes2008enhancing}, the reweighted $\ell_{1}$ method initializes all the penalties on different weights to one. In our problem, since we have pretrained models, we initialize $P_i$ using the parameters $W_i$ in the pretrained model. We use SGD or ADAM \cite{kingma2014adam} to solve the regularization problem (\ref{prob_general}). We set the parameters of the pretrained model as the starting point of the first iteration of the reweighted method, and we set the solution obtained after one iteration of the reweighted method as the starting point of the next iteration. The above approaches we used (the way to initialize $P_i$ and set starting point) can reduce the total computational time in the reweighted method. After using the reweighted method, we remove the weights (or group of weights) that are close to zero and retrain the DNN using the remaining non-zero weights. Algorithm~\ref{alg:example2} summarizes a single step of our proposed method.

Then we mask the gradients of the weights we already set to zeros (these zero weights will no longer change), and use reweighted method to generate further sparsity based on the model found in the first step. We observe that after the first step, we obtain a sparse model with sparsity larger than state-of-the-art works \cite{zhang2018systematic,ren2019admm,ye2019progressive}. However, we can further increase the degree of sparsity by using the reweighted method repeatedly. 

Since the loss function in the regularization problem is non-convex, we cannot find the globally optimum of this problem. This impacts the performance of the reweighted method to search for a model with high degree of sparsity in a single step.
However, if we use the single step repeatedly, we can keep the balance between the degree of sparsity and accuracy of the model in each step. Then we can finally achieve a highly sparse model without  accuracy loss. More specifically, 
we use a moderate $\lambda$ and apply the reweighted method for several steps, and thus obtain a highly sparse model with the competitive accuracy.



\section{An Integrated Framework for the Combination of Different DNN Weight Compression Tasks}

\subsection{Motivation}
Besides the non-structured and structured pruning, several works on the combination of different DNN weight compression tasks have achieved great performance on reducing the storage and computation of DNNs. Some combinations have different intrinsic properties. For example, pattern pruning imposes a hard constraint while kernel-wise pruning does not. Therefore, it is difficult for the kernel-wise pruning to specify the sparsity level in each layer if one wants to use a hard constraint. When combining these two pruning tasks, it is more efficient to apply our proposed reweighted methods to the kernel-wise pruning and formulate the pattern pruning as a hard constraint for the ADMM to solve. Similar methods of the integrated framework can be applied to non-structured weight pruning and weight quantization.



\subsection{Problem Formulations and Solutions}
The above challenges motivate us to present an integrated framework for the combination of different DNN weight compression tasks, which is
\begin{equation}
\label{opt_mix}
\begin{aligned}
& \underset{ \{{{W}}_{i}\},\{{{b}}_{i} \}}{\text{minimize}}
& & f \big( \{{{W}}_{i}\}_{i=1}^N, \{{{b}}_{i} \}_{i=1}^N \big)+ \lambda\sum_{i=1}^{N} R({{P}}_{i}^{(l)},{{W}}_{i}),
\\ & \text{subject to}
& & {{W}}_{i}\in {{S}}_{i}, \; i = 1, \ldots, N,
\end{aligned}
\end{equation}
where $\lambda$ is the penalty parameter, $R({{P}}_{i}^{(l)},{{W}}_{i})$ is the sparsity or group sparsity regularization term of our reweighted method, $S_i$ is the constraint set on the weights and it can be set differently according to different compression tasks. In our framework we can handle the hard constraints of one task and search the non-critical weights or structures of the other task.

The above problem can be equivalently rewritten as,
\begin{equation}
\begin{aligned}
\label{no_constraint}
 \underset{ \{{{W}}_{i}\},\{{{b}}_{i} \}}{\text{minimize}} ~~~~
 f \big( \{{{W}}_{i}\}_{i=1}^N, \{{{b}}_{i} \}_{i=1}^N \big) &+ \lambda\sum_{i=1}^{N} R({{P}}_{i}^{(l)},{{W}}_{i})\\ &+ \sum_{i=1}^{N} g_i({{W}}_{i}),
\end{aligned}
\end{equation}
where
\begin{eqnarray*}g_{i}({{W}}_{i})=
\begin{cases}
 0 & \text { if } {{W}}_{i}\in {{S}}_{i}, \\ 
 +\infty & \text { otherwise. }
\end{cases}
\end{eqnarray*}

Problem (\ref{no_constraint}) can be equivalently rewritten as
\begin{equation*}
\begin{aligned}
& \underset{ \{{{W}}_{i}\},\{{{b}}_{i} \}}{\text{minimize}}
& & f \big( \{{{W}}_{i}\}_{i=1}^N, \{{{b}}_{i} \}_{i=1}^N \big)+ \lambda\sum_{i=1}^{N} R({{P}}_{i}^{(l)},{{W}}_{i}) \\& &&  + \sum_{i=1}^{N}  g_i({{Z}}_{i}),
\\ & \text{subject to}
& & {{W}}_{i}={{Z}}_{i}, \; i = 1, \ldots, N.
\end{aligned}
\end{equation*}

To solve the problem above, we decompose it into two sub-problems. 
The first one is
\begin{equation}
\begin{aligned}
\label{subproblem_1}
 \underset{ \{{{W}}_{i}\},\{{{b}}_{i} \}}{\text{minimize}}
\ \ \ f \big( \{{{W}}_{i} \}_{i=1}^N, \{{{b}}_{i} \}_{i=1}^N \big) &+\lambda\sum_{i=1}^{N} R({{P}}_{i}^{(l)},{{W}}_{i})\\ &+\sum_{i=1}^{M} \frac{\rho}{2}  \| {{W}}_{i}-{{Z}}_{i}^{k}+{{U}}_{i}^{k} \|_{F}^{2},
\end{aligned}
\end{equation}
where $\rho$ is the penalty parameter in the ADMM, and ${{U}}_{i}^{k}$ is the dual variable. Problem (\ref{subproblem_1}) can be solved by SGD or Adam.

The second sub-problem is given by
\begin{equation}
\label{subproblem_2}
 \underset{ \{{{Z}}_{i} \}}{\text{minimize}}
\ \ \ \sum_{i=1}^{M} g_{i}({{Z}}_{i})+\sum_{i=1}^{M} \frac{\rho}{2} \| {{W}}_{i}^{k+1}-{{Z}}_{i}+{{U}}_{i}^{k} \|_{F}^{2}. \\
\end{equation}
The closed-form solution of the above problem is given in \cite{boyd2011distributed}:
\begin{equation}
  {{Z}}_{i}^{k+1} = {{{\Pi}}_{{{S}}_{i}}}({{W}}_{i}^{k+1}+{{U}}_{i}^{k}),
\end{equation}
where ${{{\Pi}}_{{{S}}_{i}}}(\cdot)$ is the Euclidean projection onto the set ${{S}}_{i}$.

Finally we update the dual variable ${{U}}_{i}^{k}$ according to
\begin{equation}
\label{dual_update}
{{U}}_{i}^{k}:={{U}}_{i}^{k-1}+{{W}}_{i}^{k}-{{Z}}_{i}^{k}. 
\end{equation}

The updates in (\ref{subproblem_1}), (\ref{subproblem_2}) and (\ref{dual_update}) complete one iteration of ADMM. We observe that it usually takes 9-12 iterations for the ADMM to converge. The regularization term of non-structured pruning and kernel-wise pruning has been discussed in Section~\ref{sec:framework} of this paper, and the hard constraint set $S_i$ of pattern pruning and weight quantization has been defined in \cite{ma2019pconv,ren2019admm}.

\section{Experiment Results for Non-structured and Structured Pruning}
\label{Sec_exp}

In this section, we evaluate our proposed framework for both non-structured pruning and structured pruning on different DNN models. We implement our non-structured pruning method on LeNet-5 \cite{lecun1998gradient} and AlexNet \cite{krizhevsky2012imagenet} models for MNIST and ImageNet ILSVRC-2012 datasets, respectively. We also implement our structured pruning method on ResNet-18 and ResNet-50 \cite{he2016deep} models for ImageNet dataset, and MobileNet-V2-1.0 \cite{sandler2018mobilenetv2} model for CIFAR-10 dataset. In every step of our framework, we use three to four iterations of the reweighted method and every iteration contains 25 epochs of DNN training. On average we need 85 epochs in every step for our method. This is lower than 150 epochs needed in the ADMM-based method. Also we achieve better performance than ADMM for both one-step pruning and multi-step pruning.

For the penalty parameter $\lambda$, it is used for adjusting the relative importance of accuracy and sparsity. Excessively small $\lambda$ fails to regularize enough non-critical weights to values close to zero and excessively large $\lambda$ is fails to minimize the loss function and the model accuracy cannot be maintained. In our experiment {we find an appropriate way to} tune $\lambda$. 
In the regularization problems (\ref{prob_general}), when we adjust $\lambda$ to set the value of the regularization term in the range of
\[
4l \le \lambda\sum_{i=1}^{N} R({{P}}_{i}^{(1)},{{W}}_{i}^{(0)}) \le 8l,
\]
we can achieve good pruning rate without accuracy loss. Where $W_i^{(0)}$ is the weights in pretrained model, and $P_{i}^{(1)}$ is derived by $W_i^{(0)}$, and $l$ is the training loss of the pretrained model.

\subsection{Non-structured pruning on LeNet-5 and AlexNet }
\subsubsection{LeNet-5 on MNIST.} Table \ref{table:LeNet-5} shows the non-structured pruning results. 
We achieve 630$\times$ pruning rate with 99.0\% accuracy and 301$\times$ pruning rate with 99.2\% accuracy. At the 99.2\% accuracy level, we achieve at least 1.5$\times$ more pruning rate compared with all the state-of-the-art. At the 99.0\% accuracy level, we achieve at at least 2.6$\times$ more pruning rate compared with all the state-of-the-art.

\begin{table}[tb]
\centering
\caption{Comparisons of overall non-structured weight pruning results on LeNet-5 for MNIST dataset.}\label{table:LeNet-5}
\begin{tabular}{p{3.5cm}p{1.5cm}p{2.0cm}}
\hline
Method & Accuracy &  Overall pruning rate \\ 
\hline
Uncompressed & 99.2\% &  1.0$\times$ \\ \hline
 {Iterative Pruning \cite{han2015learning}}  & 99.2\% &  12.5$\times$ \\ \hline
 {One-step ADMM \cite{zhang2018systematic}}  & 99.2\% &  71.2$\times$ \\ \hline
 Hoyer-Square~\cite{Yang2020DeepHoyer} &  99.2\% & 122$\times$ \\ \hline
 Progressive ADMM \cite{ye2019progressive} & 99.2\% &  200$\times$ \\ \hline 
 Progressive ADMM \cite{ye2019progressive} & 99.0\% &  246$\times$ \\ \hline
\bf{Our method (one-step)}  & \textbf{99.2\%} &  \bf{301$\times$} \\ \hline 
\bf{Our method}  & \textbf{99.0\%} &  \bf{630$\times$} \\ \hline
\end{tabular}
\end{table}



\subsubsection{AlexNet on ImageNet.} Table~\ref{table:AlexNet1} shows the non-structured pruning results. 
The top-5 accuracy of the baseline model we use is 82.4\%. 
In order to highlight the difference of the obtained accuracy by using different pruning methods, we use the relative accuracy loss against the baseline accuracy of each method. 
Note that the pruning rates of early works are less than or around 20$\times$. A recent work \cite{ye2019progressive} achieves 36$\times$ pruning rate with 82.0\% top-5 accuracy. We use the same baseline as \cite{ye2019progressive} and achieves 45$\times$ pruning rate with 82.0\% top-5 accuracy.

\begin{table}[tb]
\centering
\caption{Comparisons of overall non-structured weight pruning results on AlexNet model for ImageNet dataset.}\label{table:AlexNet1}
\begin{tabular}{p{3.1cm}p{1.5cm}p{1.5cm}p{1cm}}
\hline
Method &  Top-5 accuracy & Relative accuracy loss  & Overall pruning rate  \\ \hline
Uncompressed & 80.2\%/82.4\%  & 0.0\% & 1$\times$ \\ \hline
Iterative Pruning \cite{han2015learning} & $80.3$\%  & $-0.1$\% & 9.1$\times$ \\ \hline
Optimal Brain Surgery \cite{guo2016dynamic} & $80.0\%$ & $0.2$\% & 17.7$\times$ \\ \hline
Hoyer-Square~\cite{Yang2020DeepHoyer} & 80.2\%  & 0.0\%  & 21.3$\times$ \\ \hline
One-step ADMM \cite{zhang2018systematic} & $80.2\%$ & $0.0$\% & 21$\times$ \\ \hline
Progressive ADMM \cite{ye2019progressive} & $82.0\%$ & $0.4$\% & 36$\times$ \\ \hline
\bf{Our method (one-step)} & \textbf{82.0\%} & \textbf{0.4\%} & \bf{37$\times$} \\ \hline
\bf{Our method} & \textbf{82.3\%} & \textbf{0.1\%} & \bf{40$\times$} \\ \hline
\bf{Our method} & \textbf{82.0\%} & \textbf{0.4\%} & \bf{45$\times$} \\ \hline
\end{tabular}
\end{table}

\subsection{Structured pruning on ResNet-18, ResNet-50 and MobileNet-V2-1.0}


\subsubsection{ResNet-18 on ImageNet.} Table \ref{table:result_img_res18} shows the structured pruning results. 
The top-5 accuracy of the baseline we use is 89.0\%. DCP method \cite{zhuang2018discrimination} only achieves 1.5$\times$ pruning rate without accuracy loss, and Struct-ADMM method \cite{zhang2018adam} achieves 2.0$\times$ and 3.0$\times$ pruning rate with 0.2\% and 1.1\% accuracy loss, respectively. In our proposed method, we can achieve 3.0$\times$ pruning rate without accuracy loss. When 0.3\% accuracy loss is tolerable, we can achieve 3.8$\times$ pruning rate for shape-wise sparsity. We also implement filter-wise sparsity together with shape-wise sparsity on ResNet-18, and totally achieve 4.2$\times$ pruning rate with 0.5\% accuracy loss.

\begin{table}[tb]
    \centering
    \caption{Structured pruning results on ResNet-18 for ImageNet dataset.}
        \begin{tabular}{p{2.3cm}p{2cm}p{1.5cm}p{1.3cm}}
            \hline
            Method & Sparsity type & Conv. pruning rate & Top-5 Accuracy \\ \hline
            Original & N/A &1.0$\times$ & 89.0\% \\ \hline
            DCP \cite{zhuang2018discrimination} & filter $\&$ shape & 1.5$\times$ & 89.0\% \\ \hline
            Struct-ADMM \cite{zhang2018adam} & shape & 2.0$\times$ & 88.8\% \\ \hline
            Struct-ADMM \cite{zhang2018adam} & shape & 3.0$\times$ & 87.9\% \\ \hline
            \textbf{Our method} & shape & \bf{3.0$\times$} & 89.0\% \\ \hline
            \textbf{Our method} & shape & \bf{3.8$\times$} & 88.7\% \\ \hline
            \textbf{Our method} & filter \& shape & \bf{4.2$\times$} & 88.5\% \\ \hline
            
            \hline
					 
        \end{tabular}
    \label{table:result_img_res18}
\end{table}

\subsubsection{ResNet-50 on ImageNet.} Table~\ref{table:result_img_res50} shows the structured pruning results. 
The top-5 accuracy of the baseline we use is 92.7\%. The early work ThiNet \cite{luo2017thinet} achieves 2.0$\times$ pruning rate with high accuracy loss, the recent works DCP \cite{zhuang2018discrimination} and Struct-ADMM \cite{zhang2018adam} achieve 2.0$\times$ pruning rate with minor or no accuracy loss. In our proposed method, we achieve 3.2$\times$ pruning rate with 92.1$\%$ Top-5 accuracy, the pruning rate is much higher than prior works.

\begin{table}[tb]
    \centering
    \caption{Structured pruning results on ResNet-50 for ImageNet dataset.}
        \begin{tabular}{p{2.3cm}p{2cm}p{1.5cm}p{1.3cm}}
            \hline
            Method & Sparsity type & Conv. pruning rate & Top-5 Accuracy \\ \hline
            Original & N/A &1.0$\times$ & 92.7\% \\ \hline
            ThiNet \cite{luo2017thinet} & shape & 2.0$\times$ & 90.0\% \\ \hline
            Struct-ADMM \cite{zhang2018adam} & shape & 2.0$\times$ & 92.7\% \\ \hline
            DCP            \cite{zhuang2018discrimination} & filter $\&$ shape & 2.0$\times$ & 92.3\% \\ \hline

            \textbf{Our method} & filter \& shape & \bf{3.2$\times$} & 92.1\% \\ \hline
            
            \hline
					 
        \end{tabular}
    \label{table:result_img_res50}
\end{table}

\begin{table}[ht]
    \centering
    \caption{Structured pruning results on MobileNet-V2-1.0 for CIFAR-10 dataset}
        \begin{tabular}{p{2cm}p{2cm}p{1.5cm}p{1.3cm}}
            \hline
            Method & Sparsity type & Conv. pruning rate & Accuracy \\ \hline
            Original & N/A  &1.0$\times$ & 94.5\% \\ \hline
            DCP \cite{zhuang2018discrimination} & filter $\&$ shape & 1.4$\times$ & 94.7\% \\ \hline
            \textbf{Our method} & filter \& shape & \bf{7.2$\times$} & 94.6\% \\ \hline
					 
        \end{tabular}
    \label{table:MBNT_cifar_str}
\end{table}


            
					 

\begin{table*}[htb]
\centering
\caption{Pattern pruning $\&$ kernel-wise pruning on VGG-16 for CIFAR-10 dataset.}\label{Table:vgg16_cifar}
\begin{tabular}{p{3.5cm}p{1.5cm}p{1.43cm}p{1.5cm}p{1.7cm}p{1.5cm}}
\hline
Method &  Accuracy & Number of pattern   & Pattern prun. rate  & Kernel-wise prun. rate & Conv. prun. rate   \\ 
\hline
PCONV (ADMM) \cite{ma2019pconv} & 93.7\% &  8   & 2.25$\times$ & 8.8$\times$ & 19.8$\times$   \\
\hline
\bf{Our method} & 93.7\% & 8  & 2.25$\times$ & 15.5$\times$ & 34.9$\times$ \\
\hline

\end{tabular}
\end{table*}

\begin{table*}[htb]
\centering
\caption{Pattern pruning $\&$ kernel-wise pruning on VGG-16 for ImageNet dataset.}\label{Table:vgg16_imagenet}
\begin{tabular}{p{3.5cm}p{1.35cm}p{1.43cm}p{1.5cm}p{1.7cm}p{1.5cm}}
\hline
Method &  Top-5 Accuracy & Number of pattern   & Pattern prun. rate  & Kernel-wise prun. rate & Conv. prun. rate   \\ 
\hline
PCONV (ADMM) \cite{ma2019pconv} & 91.6\% &  8   & 2.25$\times$ & 3.1$\times$ & 7.0$\times$   \\
\hline
\bf{Our method} & 91.6\% & 8  & 2.25$\times$ & 5.8$\times$ & 13.1$\times$ \\
\hline

\end{tabular}
\end{table*}

\begin{table*}[htb]
\centering
\caption{Model size compression rate on LeNet-5 for MNIST dataset.}\label{Table:LeNetQuantization}
\begin{tabular}{p{3cm}p{1.3cm}p{1.2cm}p{1cm}p{1cm}p{1.6cm}p{2.6cm}}
\hline
Method &  Accuracy loss  & Pruning rate  & Conv. quant. & FC quant. & Data Comp. rate & Model Comp. rate  (including
index) \\ 
\hline
Original LeNet-5 & 0.0\% &  1.0$\times$   & 32b & 32b & 1.0$\times$  & 1.0$\times$ \\
\hline
Iterative pruning \cite{han2015deep} & 0.1\% & 12$\times$  & 8b & 5b & 70.2$\times$ & 33$\times$\\
\hline
ADMM-NN\cite{ren2019admm} & 0.2\% & 167$\times$  & 3b & 2b & 1910$\times$ & 623$\times$ \\
\hline
\bf{Our method} & 0.2\% & 385$\times$  & 3b & 2b & 4403$\times$ & {1014$\times$} \\
\hline

\end{tabular}
\end{table*}

\begin{table*}[htb]
\centering
\caption{Model size com ressionrate pon AlexNet for ImageNet dataset.}\label{Table:AlexNetQuantization}
\begin{tabular}{p{3cm}p{1.3cm}p{1.2cm}p{1cm}p{1cm}p{1.6cm}p{2.6cm}}
\hline
Method &  Accuracy loss  & Pruning rate  & Conv. quant. & FC quant. & Data Comp. rate & Model Comp. rate (including
index) \\ 
\hline
Original AlexNet & 0.0\% &  1.0$\times$   & 32b & 32b & 1.0$\times$  & 1.0$\times$ \\
\hline
Iterative pruning \cite{han2015deep} & 0.0\% & 9.1$\times$  & 8b & 5b & 45$\times$ & 27$\times$\\
\hline
ADMM-NN\cite{ren2019admm} & 0.2\% & 27$\times$  & 5b & 3b & 231$\times$ & 99$\times$ \\
\hline
\bf{Our method} & 0.2\% & 34$\times$  & 5b & 3b & 291$\times$ & {115$\times$} \\
\hline

\end{tabular}
\end{table*}

\subsubsection{MobileNet-V2-1.0 on CIFAR-10.} Table~\ref{table:MBNT_cifar_str} shows the structured pruning results. 
We use our pretrained baseline model with an accuracy of 94.50\%. Compared with DCP~\cite{zhuang2018discrimination} which is the state-of-the-art MobileNet-V2-1.0 pruning technique, our method significantly outperform it. We achieve 7.2$\times$ pruning rate without accuracy loss. 

Overall, using the same training trails, our method can achieve higher pruning rate than the prior works. For small to large-scale dataset, our proposed method significantly outperforms the state-of-the-art in terms of pruning rate and accuracy, leading to light weight storage and computation. 


\section{Experiment Results for the Combination of Different DNN Weight Compression Tasks}
In this section, we evaluate the performance of our proposed integrated framework for the combination of different DNN weight compression tasks. We apply our method on the combination of pattern and kernel-wise pruning on VGG-16 \cite{simonyan2014} for CIFAR-10 and ImageNet dataset. We also apply our method on the combination of non-structured pruning and weight quantization, e.g., on LeNet-5 and AlexNet for MNIST and ImageNet dataset, respectively.

\subsection{Combination of Pattern and Kernel-wise Pruning}
We compare our integrated method on the combination of pattern and kernel-wise pruning with PCONV \cite{ma2019pconv}, which used the ADMM-based hard constraint method only, the results are shown in Table \ref{Table:vgg16_cifar} and Table \ref{Table:vgg16_imagenet}. For the pattern pruning, we use the same hard constraint as PCONV, which requires the structure of every $3\times3$ kernel to belong to the shapes of 8 different patterns with 4 non-zero elements, thus the pruning rate is fixed to be 9/4=2.25. For the kernel-wise pruning, different from the hard constraint used in PCONV, we use our reweighted regularization method and achieve higher pruning rate. In conclusion,  we achieve 34.9$\times$ pruning rate on the convolutional layers of VGG-16 on CIFAR-10 dataset. This is 1.76 times more than PCONV (19.8$\times$ pruning rate). We also achieve 13.1$\times$ pruning rate on the convolutional layers of VGG-16 for ImageNet dataset. This is 1.87 times more than PCONV (7.0$\times$ pruning rate).

\subsection{Combination of Non-structured Pruning and Quantization}
We compare our integrated method on the combination of non-structured pruning and weight quantization with iterative pruning \cite{han2015deep} and ADMM-NN \cite{ren2019admm}, the results are shown in Table \ref{Table:LeNetQuantization} and Table \ref{Table:AlexNetQuantization}. When combining weight pruning with low bits weight quantization, we cannot achieve extremely high pruning rate. However, the total compression rate is higher than employing weight pruning only. We achieve 1014$\times$ compression rate on LeNet-5 for MNIST dataset and 115$\times$ compression rate on AlexNet for ImageNet dataset, which is respectively higher than 623$\times$ and 99$\times$ compression rate achieved by ADMM-NN.

\section{Conclusion}
\label{sec:conclusion}
In this paper, we propose a unified DNN weight pruning framework with dynamically updated regularization terms bounded by the designated constraint, which can generate both non-structured sparsity and different kinds of structured sparsity.  In our proposed framework, we first use reweighted method to regularize the model, then remove the weights which are close to zero and mask the gradient of these weights to ensure that they no longer update, and we retrain the remaining non-zero weights to retrieve the accuracy. We also propose an integrated framework to combine different pruning tasks, such as pattern pruning and kernel-wise pruning. Experimental results show that we achieve higher pruning rate than state-of-the-art for both non-structured, structured, and combined pruning with negligible accuracy degradation.

\section{Acknowledgments}
{This research was supported by the National Science Foundation under awards CAREER CMMI-1750531, ECCS-1609916, CCF-1919117, CNS-1909172 and CNS-1739748.}



\end{document}